\documentclass{article}
\usepackage{spconf,amsmath,graphicx,hyperref}
\usepackage{xcolor}
\usepackage{booktabs}
\usepackage{float}

\title{Ara-Best-RQ: Multi Dialectal Arabic SSL}

\name{Haroun Elleuch$^{1,2}$, Ryan Whetten$^2$, Salima Mdhaffar$^2$, Yannick Estève$^2$, Fethi Bougares$^{1,2}$\thanks{This work was granted access to the HPC resources of IDRIS under the allocations AD011015051, AD011012551R3, AD011012108R, AD011014732 and A0131013821 made by GENCI.}}

\address{
$^1$ELYADATA, Paris, France \\
$^2$Laboratoire Informatique d'Avignon, Avignon Université, Avignon, France \\
}

\begin{document}
\ninept
\maketitle
\begin{abstract}
We present Ara-BEST-RQ, a family of self-supervised learning (SSL) models specifically designed for multi-dialectal Arabic speech processing. Leveraging 5,640 hours of crawled Creative Commons speech and combining it with publicly available datasets, we pre-train conformer-based BEST-RQ models up to 600M parameters. Our models are evaluated on dialect identification (DID) and automatic speech recognition (ASR) tasks, achieving state-of-the-art performance on the former while using fewer parameters than competing models. We demonstrate that family-targeted pre-training on Arabic dialects significantly improves downstream performance compared to multilingual or monolingual models trained on non-Arabic data. All models, code, and pre-processed datasets will be publicly released to support reproducibility and further research in Arabic speech technologies.
\end{abstract}


%
\section{Introduction} \label{sec:intro}
Speech self-supervised learning (SSL) has emerged as a powerful paradigm for speech processing tasks such as automatic speech recognition (ASR). Unlike traditional approaches that rely on costly annotated datasets, SSL leverages large amounts of unlabeled data to learn high-quality general-purpose representations. This has been shown to improve transferability across tasks and languages. Models such as wav2vec 2.0~\cite{baevski2020wav2vec} and BEST-RQ~\cite{chiu2022self} have demonstrated remarkable success, and their multilingual variants (e.g. XLS-R~\cite{babu2021xls}, w2v-BERT 2.0~\cite{loic2023seamless}, Google USM~\cite{zhang2023google}) pave the way for universal speech models. SSL is therefore especially promising in low-resource settings, where labeled data is scarce.

Despite these advances, Arabic speech remains underrepresented in SSL research. While multilingual SSL models include some Arabic, training data is still dominated by English and other high-resource languages. Moreover, the Arabic content in these models and datasets is primarily Modern Standard Arabic (MSA)~\cite{babu2021xls, loic2023seamless}, with only a few studies considering dialectal speech in bespoke Arabic-focused models~\cite{djanibekov-etal-2025-dialectal}. This imbalance poses significant challenges for Arabic, particularly its dialects, which vary widely in phonology, vocabulary, and usage.
Another factor hindering progress in Dialectal Arabic SSL is the absence of publicly available speech collections suitable for SSL models, which require a vast amount of data for pre-training~\cite{loic2023seamless, hsu2021hubert}.

When it comes to mitigating such gaps with low-resource languages, researchers have explored language- or family-specific SSL pre-training. For example, LeBenchmark~\cite{evain2021lebenchmark} and Pantagruel~\cite{le2026pantagruel} provide SSL models for French, while AfriHuBERT~\cite{alabi2024afrihubert} targets African languages, both showing that focused pre-training yields better performance than multilingual models that underrepresent target languages. However, existing initiatives still leave a substantial gap for Arabic: dialectal diversity is not adequately captured in existing datasets, and models trained on these datasets fail to generalize across its many varieties.
Most existing efforts have focused on benchmarking multilingual models on Arabic~\cite{mdhaffar2024performance}, rather than on building resources and models that explicitly account for its dialectal variation.

In this work, we address this gap by building the first large-scale multi-dialectal Arabic SSL resource and models. Our contributions are threefold:

\begin{itemize}
    \item Models: We train and open-source a family of SSL models, Ara-BEST-RQ, dedicated to Arabic and its dialects.
    \item Dataset: We curate and release 5,640 hours of Creative Commons speech data covering 20 Arabic dialects, which is, to the best of our knowledge the largest collection of Arabic speech to date.
    \item Evaluation: We provide a preliminary study demonstrating strong results in dialect identification (DID) and ASR, setting a new state-of-the-art in the former and achieving comparable performance against other state-of-the-art SSL models.
\end{itemize}
We release the Ara-BEST-RQ models, code, and the crawled dataset at the following link: \url{https://github.com/elyadata/Ara-BEST-RQ}.

\section{Related Work}

Recent years have seen increasing efforts to develop self-supervised learning (SSL) models for Arabic speech. One notable example is ArTST \cite{toyin-etal-2023-artst}, which builds on the SpeechT5 architecture \cite{ao2021speecht5} and is designed for both speech-to-text (ASR) and text-to-speech (TTS) tasks. The authors show that fine-tuning a model pre-trained on English alone is not competitive with pre-training a model specifically on Arabic. However, ArTST has several limitations: it does not support dialectal Arabic, its pre-training is restricted to a single, predominantly MSA-based dataset (MGB-2), and it relies on an English-only ASR encoder (HuBERT) to generate targets for speech pre-training.

The later ArTST-v2 \cite{djanibekov-etal-2025-dialectal} incorporates dialectal datasets into the pre-training process, demonstrating improved ASR performance in both supervised fine-tuning and zero-shot settings. Nevertheless, the overall scale of the model and the combined dataset remain relatively small. Another recent approach, Aswat \cite{alkanhal-etal-2023-aswat}, also leverages SSL speech encoders such as wav2vec 2.0 \cite{baevski2020wav2vec} and data2vec \cite{baevski2022data2vec}. These systems are mostly pre-trained on MSA speech (MGB-2, Common Voice) in addition to their Aswat dataset, which is largely MSA as well. However, neither the datasets nor the models are publicly available, and the evaluations focuses primarily on MSA ASR, rather than employing multiple downstream tasks or dialectal evaluation. Similar to ArTST, the dataset and model scale remain limited.

In contrast, our approach leverages the BEST-RQ architecture \cite{chiu2022self, whetten2024open} with a conformer-based speech encoder, without pre-training a text decoder. It is trained on up to 14k hours of Arabic and multilingual speech—substantially larger than the resources used in ~\cite{toyin-etal-2023-artst, djanibekov-etal-2025-dialectal, alkanhal-etal-2023-aswat}. This enables us to support multiple downstream tasks across various Arabic dialects.


\section{Dataset} \label{sec:dataset}

In this work, we assemble two datasets: a crawled dataset and a dataset that combines our crawled data with other publicly available datasets.

\begin{table}[t]
\centering
\begin{tabular}{l r r}
\toprule
\textbf{Statistic} & \textbf{Crawled dataset} & \textbf{Combined dataset} \\
\midrule
Total duration            & 5,639h 04m 27s & 13,723h 08m 43s \\
Minimum duration          & 1.00 s         & 1.0 s \\
Maximum duration          & 20.0 s         & 20.0 s \\
Mean duration             & 4.97 s         & 5.30 s \\
Standard deviation        & 5.10 s         & 4.60 s \\
25th percentile (Q25)     & 1.72 s         & 2.36 s \\
Median                    & 2.91 s         & 3.93 s \\
75th percentile (Q75)     & 5.66 s         & 6.00 s \\
80th percentile (Q80)     & 6.91 s         & 7.00 s \\
90th percentile (Q90)     & 12.70 s        & 11.12 s \\
99th percentile (Q99)     & 20.0 s         & 20.0 s \\
\bottomrule
\end{tabular}
\caption{Comparison of segment duration statistics between the crawled dataset and the combined dataset.}
\label{tab:duration_comparison}
\end{table}

\subsection{Crawled Dataset}
We crawled more than 35,000 Creative Commons video links from YouTube from approximately 8800 channels. All the links were subsequently inspected to filter out offensive content. We did not use the geotags provided by YouTube to source the dialect metadata, as we found them to be consistently unreliable. The remaining 26k videos were downloaded, and their raw audio was converted to mono PCM at 16 kHz. Speech segments were extracted using the Silero~\cite{silero} voice activity detection tool. Consecutive detected segments that were temporally close, within 250 milliseconds, were merged. Segments longer than 20 seconds were split, and segments shorter than 1 second were discarded, resulting in a total of 3.86M speech segments amounting to 5640 hours. 
For efficient I/O during pre-training, all audio files were organized according to these speech boundaries.



\subsection{Combined Dataset}
We sourced most large- and small-scale publicly available datasets to combine a large pre-training dataset. After removing overlapped content from the datasets and discarding segments shorter than 1 second, we obtain a combined duration of 13723 hours, including our crawled dataset. 

\begin{figure}[!htb]
    \centering
    \includegraphics[width=0.8\linewidth]{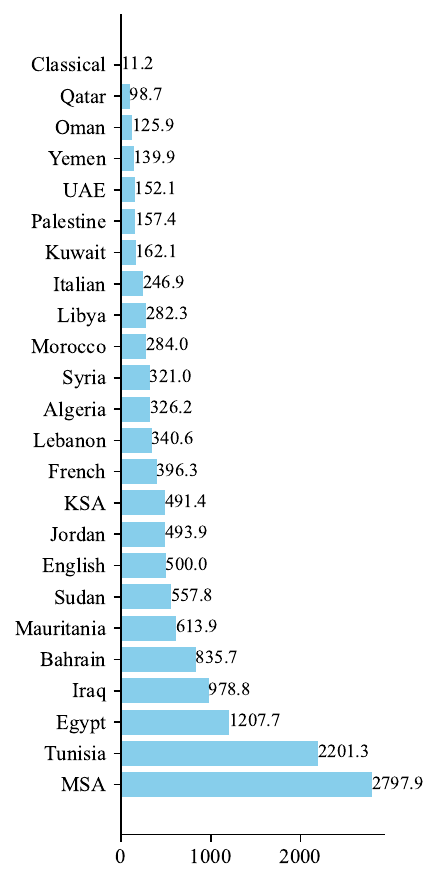}
    \caption{Distribution of the full training set (in hours) by dialect.}
    \label{fig:dialect_dist}
\end{figure}

In addition to Modern Standard Arabic (MSA) and Dialectal Arabic (DA), the dataset also includes Classical Arabic from ClArTTS~\cite{kulkarni2023clartts} in addition to Italian, French, and English from CommonVoice 16.1~\cite{ardila-etal-2020-common}. Only 500 hours of English and 396 hours of French were sampled to avoid over-representation. The sampling was performed in a way to ensure gender balance using the most recent samples.
The datasets used, their durations, and language or dialects are presented in Table~\ref{tab:datasets}
A breakdown of the languages and dialects of the combined dataset is shown in Fig.~\ref{fig:dialect_dist}. Our best-performing DID model (see Section~\ref{subsec:adi-results}) was used to tag segments where the dialect information is unavailable.


\begin{table*}[!htb]
\centering
\label{tab:datasets}
\begin{tabular}{llll}
\toprule
\textbf{Dataset Name} & \textbf{MSA / Dialects} & \textbf{Duration (hours)} \\
\midrule
ADI-17 \cite{adi17}                                & 17 dialects                                   & 3,000 \\
ADI-5 \cite{adi5}                                  & 5 dialects                                     & 53.6  \\
ArabCeleb \cite{arab-celeb}                        & Tunisian                                       & 2.6   \\
Arabic Speech Commands \cite{ghandoura2021building} & MSA                                            & 3.33  \\
Arabic Speech Corpus \cite{halabi2016arabic}      & MSA                                            & 3.8   \\
Arabic SSL (Ours)                                  & Multi-dialect                                  & 5,640.93 \\
ClArTTS \cite{kulkarni2023clartts}                & Classical Arabic                               & 12    \\
CommonVoice 16.1 (AR) \cite{ardila-etal-2020-common} & MSA                                         & 156   \\
CommonVoice 16.1 (FR) \cite{ardila-etal-2020-common} & French                                      & 396   \\
CommonVoice 16.1 (EN) \cite{ardila-etal-2020-common} & English                                    & 246.9   \\
CommonVoice 16.1 (IT) \cite{ardila-etal-2020-common} & Italian                                    & 500   \\
ESCWA-CA\footnote{\url{https://arabicspeech.org/escwa}} & MSA + Code Switching                       & 7.26  \\
LinTO \cite{linagora2024Linto-tn}                 & Tunisian                                    & 93.1  \\
Massive Arabic Speech Corpus (MASC) \cite{masc-dataset} & MSA + 23 dialects                      & 1,000 \\
MediaSpeech (AR) \cite{mediaspeech2021}           & MSA                                           & 10    \\
MGB-2 \cite{mgb2}                                 & MSA                                           & 1,200 \\
MGB-3 \cite{mgb3}                                 & Egyptian                                      & 16    \\
MGB-5 \cite{mgb5}                                 & Moroccan                                      & 16    \\
Mixat \cite{al-ali-aldarmaki-2024-mixat}         & Emirati + Code Switching                     & 14.9  \\
Munazarat 1.0 \cite{khader-etal-2024-munazarat}  & MSA                                           & 67.5  \\
QASR \cite{mubarak-etal-2021-qasr}               & Multi-dialect                                 & 2,000 \\
Saudi Audio Dataset for Arabic (SADA) \cite{sada-dataset} & Saudi                                    & 667   \\
The Arabic Speech Corpus for Isolated Words \cite{isolated-words-dataset} & MSA                  & 2.25  \\
Tunisian MSA\footnote{\url{https://openslr.org/46/}} & MSA                                          & 12.6  \\
TunSwitch (Weakly labelled data only) \cite{tunswitch} & Tunisian + Code Switching                 & 158.6 \\
\bottomrule
\end{tabular}
\caption{Datasets used for pre-training.}

\end{table*}

\section{Experiments \& Results}

\begin{table*}[!htb]
\centering
\begin{tabular}{@{}l|c|cccc|c@{}}
\toprule                                 
Model                              & Size      & CV 19.0           & MGB-3          & MGB-5          & TARIC-SLU      & Average        \\ \midrule
HuBERT-large                       & 320.2 M   & 30.3             & 52.54          & 65.20          & 26.45          & 43.62          \\
XLS-R-128                          & 320.2 M   & 27.51            & 61.70          & 62.81          & 25.33          & 44.33          \\ 
Ara-BEST-RQ crawled 300M           & 311.6 M   & \textbf{18.67}            & \textbf{30.85}          & \textbf{54.18} &  \textbf{23.98} & \textbf{31.92} \\ \midrule \midrule
w2v-BERT 2.0                       & 590.0 M   & \textbf{18.56}   & \textbf{28.42} & \textbf{52.92} & 21.47          & \textbf{30.34} \\
Ara-BEST-RQ crawled 600M           & 611.3 M   & 19.50            & 30.83          & 55.78          & 22.41          & 32.13         \\
Ara-BEST-RQ combined data 600M     & 611.6 M   & 18.59            & 28.78          & 54.54          & \textbf{21.14} & 30.76          \\ \bottomrule
\end{tabular}
\caption{WER obtained on the test splits of the datasets. For MGB-3, the average WER across all annotators is reported. The "Average" column reports the average score obtained by the system across all datasets. CV 19.0 stands for the Arabic split of Common Voice 19.0.}
\label{tab:test-wer-results}
\end{table*}

\subsection{Ara-BEST-RQ  pre-training}
We adopt the BEST-RQ framework for its demonstrated efficiency and performance~\cite{chiu2022self, whetten2024open, zhang2023google}, using the Speechbrain implementation~\cite{speechbrain, whetten2024open}. Two model variants are pretrained: 300M and 600M parameters, both employing a streaming architecture with Dynamic Chunk Training. During training, audio is segmented into chunks of approximately 40 ms, with batch-wise probabilistic sampling of chunk sizes between 8 and 32 frames (probability = 0.6). Left context is also randomly limited with probability 0.75, ranging from 2 to 32 chunks, enabling the model to learn robust representations over both short and long temporal contexts.

The 300M model uses a conformer-based encoder with 24 layers, model dimension 848, 8 attention heads, and feedforward layers of dimension 2048. The 600M model increases the encoder width to 1024 and feedforward dimension to 4096, while keeping the number of layers and attention heads unchanged. Both variants employ GELU activations, layer normalization before attention, and Relative Position Multi-Head Attention to efficiently capture temporal dependencies. A convolutional front-end with two blocks preprocesses the input, preserving local spectral features. During pretraining, masking is applied with a mask length of 4 and probability 0.15 (resulting in a total mask of 60\% following~\cite{whetten2024open}), and a random projection quantizer with 4096 codebook entries of dimension 16 converts continuous representations into discrete targets.

The 300M models are pretrained using 16$\times$A100 80GB GPUs, while the 600M variants use 32$\times$H100 80GB GPUs. All models are trained with a batch duration of 450 seconds. The resulting models, dubbed Ara-BEST-RQ, are pretrained on both the crawled and combined datasets described in Section~\ref{sec:dataset}, using the combined validation splits to compute the loss.

\begin{table}[H]
\centering
\begin{tabular}{@{}llcc@{}}
\toprule
Model size & Training set  & Train loss & Valid. loss \\ \midrule
300M & Crawled  & 3.81          & 3.86              \\
300M & Combined & 6.61          & 6.10               \\ \midrule
600M & Crawled  & \textbf{3.53} & 3.70               \\
600M & Combined & 3.57          & \textbf{3.40}      \\ \bottomrule
\end{tabular}
\caption{Train and validation losses of Ara-BEST-RQ  models during pre-training.}
\label{tab:pretraining}
\end{table}

Table~\ref{tab:pretraining} presents the validation losses after 300k updates. The 300M model pretrained on the combined dataset fails to converge, likely due to its limited capacity to handle the greater variability of the larger and more diverse data. Consequently, this model is excluded from downstream evaluations. In contrast, the 600M model trained on the combined dataset reaches the lowest validation loss, whereas training the same model on the crawled dataset alone shows signs of overfitting.

\subsection{Downstream Fine-tuning}
To assess the performance of our Ara-BEST-RQ models, we fine-tuned them for the dialect identification (DID) and ASR tasks.

\subsubsection{Automatic Speech Recognition}
For ASR, we benchmark our models against three strong SSL baselines: (i) HuBERT-large-1160k~\cite{hsu2021hubert}, pretrained on LibriLight, a large-scale English-only corpus; (ii) XLS-R-128~\cite{babu2021xls}, a 300M-parameter model trained on 128 languages including Arabic; and (iii) w2v-BERT 2.0~\cite{loic2023seamless}, a 590M-parameter model pretrained on 4.5M hours of multilingual audio spanning 143 languages. Fine-tuning is performed with a three-layer feedforward network and a CTC classification head, except for w2v-BERT 2.0, where a linear probe provides better performance. All models use a shared tokenizer trained on the combined training splits of the evaluation datasets.

We evaluate on four dialectal benchmarks—MGB-3 (Egyptian), MGB-5 (Moroccan), and TARIC-SLU (Tunisian) \cite{mdhaffar2024taric}, alongside the Arabic split of Common Voice 19.0 to assess MSA performance. Table~\ref{tab:test-wer-results} reports the WERs, showing that Ara-BEST-RQ 300M outperforms the baselines of similar sizes on all the datasets. 

By up-scaling the model to 600M parameters, we cannot see the same gain in comparison to  w2v-BERT 2.0 that achieves the lowest overall average WER.
However, the 600M Ara-BEST-RQ variants are still competitive considering that w2v-BERT 2.0 was trained on 4.5M hours of multilingual audio, whereas Ara-BEST-RQ relies exclusively on 6k–14k hours of Arabic speech. 
These findings underscore the effectiveness of domain-focused pretraining and suggest that massive multilingual models are not always the best choice for specialized tasks such as Arabic ASR. 
We expect that increasing the size of the pre-training dataset would shrink the observed performance gap, which we will target in future work.


\subsubsection{Dialect Identification} \label{subsec:adi-results}

\begin{table}[H]
\begin{tabular}{@{}lcc|cc@{}}
\cmidrule(l){2-5}
                                       & \multicolumn{2}{c|}{Validation} & \multicolumn{2}{c}{Test} \\ \midrule
\multicolumn{1}{l|}{Model}             & Acc. (\%)       & F1 (\%)       & Acc. (\%)    & F1 (\%)   \\ \midrule
\multicolumn{1}{l|}{Whisper-large \cite{elleuch2025adi}} & 95.76          & 95.73          & 94.83          & 94.83          \\ 
\multicolumn{1}{l|}{w2v-BERT 2.0} & NC  & NC & NC & NC               \\ \midrule
\multicolumn{1}{l|}{Crawled 300M}                   & \textbf{97.21} & \textbf{97.17} & \textbf{96.02} & \textbf{95.98} \\
\multicolumn{1}{l|}{Crawled 600M} & 92.86           & 92.87         & 91.05        & 91.04     \\
\multicolumn{1}{l|}{Combined data 600M}     & 94.66           & 94.71         & 92.05        & 92.07     \\ \bottomrule
\end{tabular}
\caption{Accuracy and weighted F1-scores obtained on the ADI-20 benchmark with our Ara-BEST-RQ models compared to SoTA. NC: Model did not converge.}
\label{tab:adi-results}
\end{table}

For Arabic DID, we use the recently released ADI-20 benchmark~\cite{elleuch2025adi}. 
We follow the authors' recipe, using ADI-20-53h for fine-tuning, and add an attention pooling layer and a classification head to the Ara-BEST-RQ models, similarly to their Whisper-based systems. Table~\ref{tab:adi-results} shows that our Ara-BEST-RQ 0.3B trained on the crawled dataset outperforms the state-of-the-art (SoTA) results in both accuracy and F1-scores for both the test and validation splits, achieving new SoTA results while having less than half the parameters of the whisper-based system (637M).
However, the 600M variants do not perform as well, especially on the test set.
w2v-BERT 2.0 using the same recipe did not converge.

\section{Limitations}
Despite the promising results, our work presents several limitations:  
\begin{itemize}
    \item \textbf{Dataset imbalance:} Although our corpus spans more than 19 dialects in addition to MSA and Classical Arabic, the distribution remains uneven (Fig.~\ref{fig:dialect_dist}). Mitigation strategies include targeted data acquisition, which is resource-intensive, or algorithmic balancing. However, both approaches are susceptible to errors introduced by biases in the automatic DID system.  
    \item \textbf{Downstream evaluation:} The evaluation of Ara-BEST-RQ has so far focused on Arabic DID and ASR. Broader downstream tasks such as end-to-end speech translation and spoken language understanding across dialects should be investigated to provide a more comprehensive assessment.  
    \item \textbf{Model scale:} State-of-the-art SSL models increasingly exceed 1B parameters~\cite{loic2023seamless, babu2021xls, zhang2023google}. Scaling Ara-BEST-RQ to larger architectures remains unexplored, while producing smaller, efficient variants for resource-constrained settings is also an important direction.  
\end{itemize}

\section{Conclusion}
We presented Ara-BEST-RQ, a family of open-source self-supervised models pretrained on large-scale Arabic speech. Evaluations on ASR and dialect identification showed that domain-focused pretraining delivers consistent improvements over strong multilingual and monolingual baselines. Notably, the 300M model trained on 5.6k hours of crawled Arabic data outperforms HuBERT-large and XLS-R, and rivals w2v-BERT 2.0 on several tasks, despite using half the parameters and orders of magnitude less training data. These results highlight the efficiency of language-family-specific SSL pretraining for underrepresented languages.
Future work will investigate more effective scaling strategies, including larger architectures, improved data curation, and lightweight variants optimized for deployment. 
We aim to collect more Arabic data, since we expect that increasing the size of the pre-training dataset will allows us to improve the performance of our 600M parameters model.
To support ongoing research, we release the Ara-BEST-RQ models, pretraining recipes, and the crawled dataset.

\section{Acknowledgements}
This work was partially funded by the ESPERANTO project.
The ESPERANTO project has received funding from the European Union’s Horizon 2020 (H2020) research and innovation program under the Marie Skłodowska-Curie grant agreement No 101007666. This work was granted access to the HPC resources of IDRIS under the allocations 
AD011015051R1,
A0181012551, 
and AD011012108R3 
made by GENCI.


\bibliographystyle{IEEEbib}
\bibliography{strings,refs}

@INPROCEEDINGS{adi17,
  author={Shon, Suwon and others},
  booktitle={ICASSP 2020}, 
  title={ADI17: A Fine-Grained Arabic Dialect Identification Dataset}, 
  year={2020},
  volume={},
  number={},
  pages={8244-8248},
  doi={10.1109/ICASSP40776.2020.9052982}}

@article{adi5,
  title={Speech recognition challenge in the wild: Arabic MGB-3},
  author={M. Ali, Ahmed and others},
  journal={ASRU 2017},
  year={2017},
  pages={316-322},
  url={https://api.semanticscholar.org/CorpusID:215825443}
}

@InProceedings{arab-celeb,
    author="Bianco, Simone and others",
    title="ArabCeleb: Speaker Recognition in Arabic",
    booktitle="AIxIA 2021 -- Advances in Artificial Intelligence",
    year="2022",
    publisher="Springer International Publishing",
    address="Cham",
    pages="338--347",
    isbn="978-3-031-08421-8"
}

@article{ghandoura2021building,
  title={Building and benchmarking an Arabic Speech Commands dataset for small-footprint keyword spotting},
  author={Ghandoura, Abdulkader and others},
  journal={Engineering Applications of Artificial Intelligence},
  volume={102},
  pages={104267},
  year={2021},
  publisher={Elsevier}
}

@phdthesis{halabi2016arabic,
  author       = {Nawar Halabi},
  title        = {Arabic Speech Corpus},
  school       = {University of Oxford},
  year         = {2016},
}

@article{kulkarni2023clartts,
  title={Clartts: An open-source classical arabic text-to-speech corpus},
  author={Kulkarni, Ajinkya and others},
  journal={arXiv preprint arXiv:2303.00069},
  year={2023}
}

@inproceedings{ardila-etal-2020-common,
    title = "Common Voice: A Massively-Multilingual Speech Corpus",
    author = "Ardila, Rosana  and others",
    booktitle = "LREC",
    month = may,
    year = "2020",
    address = "Marseille, France",
    publisher = "ELRA",
    url = "https://aclanthology.org/2020.lrec-1.520",
    pages = "4218--4222",
    language = "English",
    ISBN = "979-10-95546-34-4",
}

@misc{linagora2024Linto-tn,
  author = {Hedi Naouara and others},
  title = {LinTO Audio and Textual Datasets to Train and Evaluate Automatic Speech Recognition in Tunisian Arabic Dialect},
  year = {2024},
  month = {October},
  note = {Good Data Workshop, AAAI 2025},
}

@INPROCEEDINGS{masc-dataset,
  author={Al-Fetyani, Mohammad and others},
  booktitle={SLT 2022}, 
  title={MASC: Massive Arabic Speech Corpus}, 
  year={2023},
  volume={},
  number={},
  pages={1006-1013},
  doi={10.1109/SLT54892.2023.10022652}}

@misc{mediaspeech2021,
      title={MediaSpeech: Multilanguage ASR Benchmark and Dataset}, 
      author={Rostislav Kolobov and others},
      year={2021},
      eprint={2103.16193},
      archivePrefix={arXiv},
      primaryClass={eess.AS}
}

@INPROCEEDINGS{mgb2,
  author={Ali, Ahmed and others},
  booktitle={SLT 2016}, 
  title={The MGB-2 challenge: Arabic multi-dialect broadcast media recognition}, 
  year={2016},
  volume={},
  number={},
  pages={279-284},

  doi={10.1109/SLT.2016.7846277}}

@INPROCEEDINGS{mgb3,
  author={Ali, Ahmed and others},
  booktitle={ASRU 2017}, 
  title={Speech recognition challenge in the wild: Arabic MGB-3}, 
  year={2017},
  volume={},
  number={},
  pages={316-322},
  doi={10.1109/ASRU.2017.8268952}}

@INPROCEEDINGS{mgb5,
  author={Ali, Ahmed and others},
  booktitle={ASRU}, 
  title={The MGB-5 Challenge: Recognition and Dialect Identification of Dialectal Arabic Speech}, 
  year={2019},
  volume={},
  number={},
  pages={1026-1033},
  doi={10.1109/ASRU46091.2019.9003960}}

@inproceedings{al-ali-aldarmaki-2024-mixat,
    title = "Mixat: A Data Set of Bilingual Emirati-{E}nglish Speech",
    author = "Al Ali, Maryam Khalifa  and
      Aldarmaki, Hanan",
    booktitle = "LREC-COLING",
    month = may,
    year = "2024",
    address = "Torino, Italia",
    publisher = "ELRA and ICCL",
    url = "https://aclanthology.org/2024.sigul-1.26",
    pages = "222--226",
}

@inproceedings{khader-etal-2024-munazarat,
    title = "Munazarat 1.0: A Corpus of {A}rabic Competitive Debates",
    author = "Khader, Mohammad M.  and others",
    booktitle = "OSACT - LREC-COLING 2024",
    month = may,
    year = "2024",
    publisher = "ELRA and ICCL",
    url = "https://aclanthology.org/2024.osact-1.3",
}

@INPROCEEDINGS{sada-dataset,
  author={Alharbi, Sadeen and others},
  booktitle={ICASSP 2024}, 
  title={SADA: Saudi Audio Dataset for Arabic}, 
  year={2024},
  volume={},
  number={},
  pages={10286-10290},
  doi={10.1109/ICASSP48485.2024.10446243}}

@inproceedings{isolated-words-dataset,
  title = {On Improving the Classification Capability of Reservoir Computing for Arabic Speech Recognition},
  booktitle = {ICANN 2014},
  author = {Alalshekmubarak, Abdulrahman and Smith, Leslie S.},
  year = {2014},
  pages = {225--232},
  publisher = {Springer International Publishing},
  address = {Cham},
  isbn = {978-3-319-11179-7}
}

@INPROCEEDINGS{tunswitch,
  author={Abdallah, Ahmed and others},
  booktitle={ICASSP 2024}, 
  title={Leveraging Data Collection and Unsupervised Learning for Code-Switched Tunisian Arabic Automatic Speech Recognition}, 
  year={2024},
  volume={},
  number={},
  pages={12607-12611},
  doi={10.1109/ICASSP48485.2024.10445734}}

@inproceedings{toyin-etal-2023-artst,
    title = "{A}r{TST}: {A}rabic Text and Speech Transformer",
    author = "Toyin, Hawau  and others",
    booktitle = "Proceedings of ArabicNLP 2023",
    address = "Singapore (Hybrid)",
    publisher = "ACL",
    pages = "41--51",
}

@inproceedings{djanibekov-etal-2025-dialectal,
    title = "Dialectal Coverage And Generalization in {A}rabic Speech Recognition",
    author = "Djanibekov, Amirbek  and others",
    booktitle = "ACL",
    year = "2025",
    url = "https://aclanthology.org/2025.acl-long.1427/",
    doi = "10.18653/v1/2025.acl-long.1427",
    pages = "29490--29502",
    ISBN = "979-8-89176-251-0"
}

@inproceedings{alkanhal-etal-2023-aswat,
    title = "Aswat: {A}rabic Audio Dataset for Automatic Speech Recognition Using Speech-Representation Learning",
    author = "Alkanhal, Lamya  and others",
    booktitle = "Proceedings of ArabicNLP 2023",
    month = dec,
    year = "2023",
    address = "Singapore (Hybrid)",
    publisher = "ACL",
    url = "https://aclanthology.org/2023.arabicnlp-1.10/",
    doi = "10.18653/v1/2023.arabicnlp-1.10",
    pages = "120--127"
}

@article{baevski2020wav2vec,
  title={wav2vec 2.0: A framework for self-supervised learning of speech representations},
  author={Baevski, Alexei and others},
  journal={NeurIPS},
  volume={33},
  pages={12449--12460},
  year={2020}
}

@inproceedings{baevski2022data2vec,
  title={Data2vec: A general framework for self-supervised learning in speech, vision and language},
  author={Baevski, Alexei and others},
  booktitle={International conference on machine learning},
  year={2022},
  organization={PMLR}
}

@article{ao2021speecht5,
  title={Speecht5: Unified-modal encoder-decoder pre-training for spoken language processing},
  author={Ao, Junyi and others},
  journal={arXiv preprint arXiv:2110.07205},
  year={2021}
}

@inproceedings{whetten2024open,
  title={Open implementation and study of best-rq for speech processing},
  author={Whetten, Ryan and others},
  booktitle={ICASSPW)},
  year={2024},
  organization={IEEE}
}

@inproceedings{chiu2022self,
  title={Self-supervised learning with random-projection quantizer for speech recognition},
  author={Chiu, Chung-Cheng and others},
  booktitle={ICML},
  pages={3915--3924},
  year={2022},
  organization={PMLR}
}

@article{hsu2021hubert,
  title={Hubert: Self-supervised speech representation learning by masked prediction of hidden units},
  author={Hsu, Wei-Ning and others},
  journal={IEEE/ACM},
  volume={29},
  pages={3451--3460},
  year={2021},
  publisher={IEEE}
}

@inproceedings{elleuch2025adi,
  title={ADI-20: Arabic Dialect Identification dataset and models},
  author={Elleuch, Haroun and others},
  booktitle={Proceedings of Interspeech},
  year={2025}
}

@misc{silero,
  author = {Silero Team},
  title = {Silero VAD: pre-trained enterprise-grade Voice Activity Detector (VAD), Number Detector and Language Classifier},
  year = {2024},
  publisher = {GitHub},
}

@article{zhang2023google,
  title={Google usm: Scaling automatic speech recognition beyond 100 languages},
  author={Zhang, Yu and others},
  journal={arXiv preprint arXiv:2303.01037},
  year={2023}
}

@article{speechbrain,
  title={Open-source conversational ai with speechbrain 1.0},
  author={Ravanelli, Mirco and others},
  journal={Journal of Machine Learning Research},
  volume={25},
  number={333},
  pages={1--11},
  year={2024}
}

@article{babu2021xls,
  title={XLS-R: Self-supervised cross-lingual speech representation learning at scale},
  author={Babu, Arun and others},
  journal={arXiv preprint arXiv:2111.09296},
  year={2021}
}

@article{loic2023seamless,
  title={Seamless: Multilingual expressive and streaming speech translation},
  author={Lo{\"\i}c, Barrault and others},
  journal={arXiv preprint arXiv: 2312.05187},
  year={2023}
}

@article{evain2021lebenchmark,
  title={Lebenchmark: A reproducible framework for assessing self-supervised representation learning from speech},
  author={Evain, Sol{\`e}ne and others},
  journal={arXiv preprint arXiv:2104.11462},
  year={2021}
}

@article{alabi2024afrihubert,
  title={AfriHuBERT: A self-supervised speech representation model for African languages},
  author={Alabi, Jesujoba O and others},
  journal={arXiv preprint arXiv:2409.20201},
  year={2024}
}

@inproceedings{mubarak-etal-2021-qasr,
    title = "{QASR}: {QCRI} Aljazeera Speech Resource A Large Scale Annotated {A}rabic Speech Corpus",
    author = "Mubarak, Hamdy  and others",
    booktitle = "ACL",
    month = aug,
    year = "2021",
    url = "https://aclanthology.org/2021.acl-long.177/",
    doi = "10.18653/v1/2021.acl-long.177",
    pages = "2274--2285",
}

@inproceedings{mdhaffar2024taric,
  title={TARIC-SLU: A Tunisian benchmark dataset for spoken language understanding},
  author={Mdhaffar, Salima and others},
  booktitle={LREC-COLING 2024},
  pages={15606--15616},
  year={2024}
}

@article{le2026pantagruel,
  title={Pantagruel: Unified Self-Supervised Encoders for French Text and Speech},
  author={Le, Phuong-Hang and others},
  journal={arXiv preprint arXiv:2601.05911},
  year={2026}
}

@inproceedings{mdhaffar2024performance,
  title={Performance Analysis of Speech Encoders for Low-Resource SLU and ASR in Tunisian Dialect},
  author={Mdhaffar, Salima and others},
  booktitle={ArabicNLP},
  pages={130--139},
  year={2024}
}

\end{document}